\documentclass{article}
\usepackage{spconf,amsmath,graphicx}
\usepackage{lineno,hyperref}
\usepackage{color}
\usepackage{subcaption}
\usepackage{array}
\usepackage{balance}
\usepackage{multirow}
\usepackage[linesnumbered]{algorithm2e}
\usepackage{caption,amssymb}
\usepackage{soul}
\usepackage{rotating}
\usepackage{enumitem}
\usepackage{flushend}

\newcommand{\WKS}[1]{\textcolor{black}{#1}}
\newcommand{\BG}[1]{\textcolor{black}{#1}}
\newcommand{\CH}[1]{\textcolor{black}{#1}}

\newenvironment{conditions}
  {\par\vspace{\abovedisplayskip}\noindent\begin{tabular}{>{$}l<{$} @{${}={}$} l}}
  {\end{tabular}\par\vspace{\belowdisplayskip}}

\title{BAITRADAR: A MULTI-MODEL CLICKBAIT DETECTION ALGORITHM\\ USING DEEP LEARNING}
\name {Bhanuka Gamage, Adnan Labib, Aisha Joomun, Chern Hong Lim, and KokSheik Wong}
\address{School of Information Technology, Monash University, 47500, Selangor, Malaysia}
\begin{document}
%
\maketitle
\begin{abstract}
Following the rising popularity of YouTube,  there is an emerging problem on this platform called \emph{clickbait}, which provokes users to click on videos using attractive titles and thumbnails. 
As a result, users ended up watching a video that does not have the content as publicized in the title. 
This issue is addressed in this study by proposing an algorithm called \emph{BaitRadar}, which uses a deep learning technique where six inference models are jointly consulted to make the final classification decision. 
These models focus on different attributes of the video, including title, comments, thumbnail, tags, video statistics and audio transcript.
The final classification is attained by computing the average of multiple models to provide a robust and accurate output even in situation where there is missing data. 
The proposed method is tested on 1,400 YouTube videos. 
On average, a test accuracy of $98\%$ is achieved with an inference time of $\leq 2s$.
\end{abstract}
\begin{keywords}
BaitRadar, 
clickbait detection, 
YouTube, 
deep learning
\end{keywords}

\section{Introduction}\label{sec:intro}
With the advancement of network infrastructures and technologies, the World Wide Web has become accessible to more people. 
Thanks to the affordable price tags of smart devices, digital contents can now be easily generated, edited and shared at one's fingertip.
\WKS{Social networking services (SNS), which link users worldwide, make it even easier to share contents. 
In fact, some SNS platforms focus on a specific type of medium, such as \emph{video} on the SNS platform called \emph{YouTube}.} 
YouTube’s market influence on the video sharing content platforms is \WKS{evidenced with} more than five hundred hours of videos uploaded every minute.
In addition, it is currently the second largest search engine and the third most visited site on the internet~\cite{Smith2019}.


Traditionally, marketers of multi-million dollar companies advertised in newspapers and local radio/television 
\WKS{broadcasting}. 
However, with the potential of reaching a much wider audience, their focus has shifted to YouTube.
As a result, the content creators on YouTube are exploiting this opportunity to make money by using the platform.
YouTube attracts content  creators to make their videos suitable for advertisements by offering money for each view, with the condition that the viewed video being compliant with their advertisement-friendly guidelines~\cite{Rosenberg2018}. 

Due to this initiative, a technique called \emph{clickbait} has emerged on YouTube.
Clickbait \WKS{exploits} an attractive title and thumbnail to incite the audience to click on a video, but the actual content does not match the published title or expectation.   
According to the information gap theory as reported by Loewenstein~\cite{Loewenstein1994}, human have the need to explore and fill the knowledge gap \WKS{between} \emph{what they know} and \emph{what they wish to know}. 
Clickbait exploits this gap by using the Title and Thumbnail. 
To uncover the gap, people will naturally click on these videos~\cite{Anand2016}. 
Solving the Clickbait problem will contribute to the reduction of time wasted online and unnecessary network traffic.
Furthermore, it creates a safer environment for children to navigate the internet in cases where clickbait leads to age-inappropriate contents.
Moveover, the recommendation engines on social media platforms can also be improved to not recommend clickbait. 
In the long run this will improve the 
\WKS{overall experience }for SNS platform users.

Considering the aforementioned issues, this study proposes a multi-model deep learning architecture to solve the clickbait problem.
Our proposed method - BaitRadar makes the following contributions: (a) combines multiple cues in a YouTube video to make inference; (b) exploits the audio transcript of the video in tandem with other cues in the video, instead of focusing only on user generated features such as comments and statistics of the video, and; (c) achieves more resilience against missing data, hence a more robust classification approach in comparison to the conventional methods.
The final classification, i.e., clickbait or non-clickbait, is made by considering the output of six models, where each model focuses on different attribute of the video. 

\section{Related Work}\label{sec::LiteratureReview}
While clickbait detection is a well researched topic in the context of web pages and articles, only a limited number of work are reported for detecting clickbait in video. 
Therefore, our literature review rely on clickbait detection in other forms such as web articles on social media. 
According to Zheng et al.~\cite{Zheng2018}, clickbait detection algorithms can be broadly divided into two main categories, namely, Lexical Similarity Algorithm (LSA), and Machine Learning Algorithms (MLA).

LSA performs well on text-based articles by focusing on the heading and the context.
It assesses how the semantics of the context and the title/headline are related. 
Interested reader may refer to 
\cite{Zheng2018,Kiros2015,Mikolov2013,Wang2011} for LSA-based clickbait detection methods that evaluate the similarity between the title and the content. 
On the other hand, MLA uses three main approaches to categorise whether a given article/video is clickbait. 
They either use hand-crafted features \cite{Potthast2016, Chakraborty2016}, user-generated features \cite{Chen2015, Zheng2017, Zannettou2018}, or automatically extracted features \cite{Anand2016, Zheng2018} to determine the patterns and important keywords in the headline and/or the content.

Based on our literature review, there is no implementation of a model with an architecture where the audio transcript or other important cues such as Title, Tags, Thumbnail, Comments, and Statistics are taken into consideration. 
Therefore, in this work, we investigate into the combination of the aforementioned cues by using a multi-model deep learning architecture to achieve better performance for clickbait detection.

\section{Proposed method - BaitRadar}\label{sec::Proposed}
Our proposed model called \emph{BaitRadar} is inspired by Zannettou et al.'s method~\cite{Zannettou2018}. 
\BG{Their approach extracts the cues from user-generated data and passes it over to a variational autoencoder.}
The key difference between our proposed method and theirs is \WKS{that} they focus on the user-generated features such as Comments and Statistics, while ours takes the audio transcript from YouTube into consideration along with Title, Thumbnail, Comments, Tags and Statistics. 

Specifically, this study implements an architecture that integrates different deep learning models, each looking into different attributes of a clickbait video. 
Here, the individual models are trained to perform on these attributes (six in total) of the video 
and provide a prediction. 
Then, the multi-model architecture fuses all prediction outcomes to infer whether the video of interest is clickbait or not. 
Specifically, the following steps are preformed to develop our architecture: 1. Dataset gathering and pre-processing; 2. Training and Evaluation of Individual Models, and; 3. Multi-model deep learning architecture.
We detail each step in the following subsections.

\subsection{Dataset Gathering and Pre-processing}
\CH{The dataset employed to train our models is generated by using the list of YouTube channels provided by Zannettou et al.~\cite{Zannettou2018}. All the details of the videos from each channel and the audio transcripts are downloaded using the YouTube API and Youtube Transcript API~\cite{youtubetranscriptapi}. 
\WKS{In total, about 14,000 videos are downloaded, where 8,591 are Clickbait videos and 5,049 are  Non-clickbait videos. 
These videos are divided into the training, validation and test sets with the proportion of 81\%, 9\%, and 10\%, respectively.}} 



\subsection{Training and Evaluation of Individual Models}
Each of the individual models from the subcategories, namely, title, audio transcript, tags, thumbnail, comments and statistics, are designed and trained to find an optimal architecture which can determine whether a video is clickbait. 
During this process, the lightest architecture of each model with the highest accuracy is investigated since our models need to have fast inference response
so that the user is notified as soon as possible when a video is classified as clickbait.
Therefore, the architecture is separated into individual training models and only the useful features are extracted from the training for integration into the combined model. 
As a result, each individual architecture is specialized to perform well for one type of feature (attribute).
For example, the text-based models utilize long short-term memory (LSTM) networks to maintain the state while the thumbnail model utilizes a convolution neural network (CNN) to extract the features in the thumbnail.
After the training is done to obtain the weights for each individual model, the models are ready to be combined.

\subsection{Multi-model Deep Learning Architecture}
Each individual model\footnote{Model parameters and hyper-parameters are listed at \url{https://baitradar.bhanukagamage.com}} is combined to build a multi-model deep learning architecture that can be utilized to classify whether the video of interest is clickbait, even when some video data or attributes are missing. 
In order to combine the individual models, a set of selected layers with the same size from all the individual models are fused by using the \CH{element wise average operation (Equation~\ref{Eq:average})}. 
This allows the final model to focus on the individual models that have higher weights in the clickbait detection process.

After combining the individual models, transfer learning is applied in the experiment by: (1) training the dense layers after the average layer 
of the combined model, and; (2) fine-tuning the entire model. 
However, \WKS{none} of the approaches yield any improvement as the gradients are stuck at local minimums. 
Therefore, the full multi-model architecture is trained from scratch in order to learn all the weights. 
Although the weights from each model does-not transfer well onto the final model, we are able to get an architecture that performs better on each attribute by training the individual model.

\begin{figure}[!t]
\centering
\includegraphics[width=1.0\linewidth]{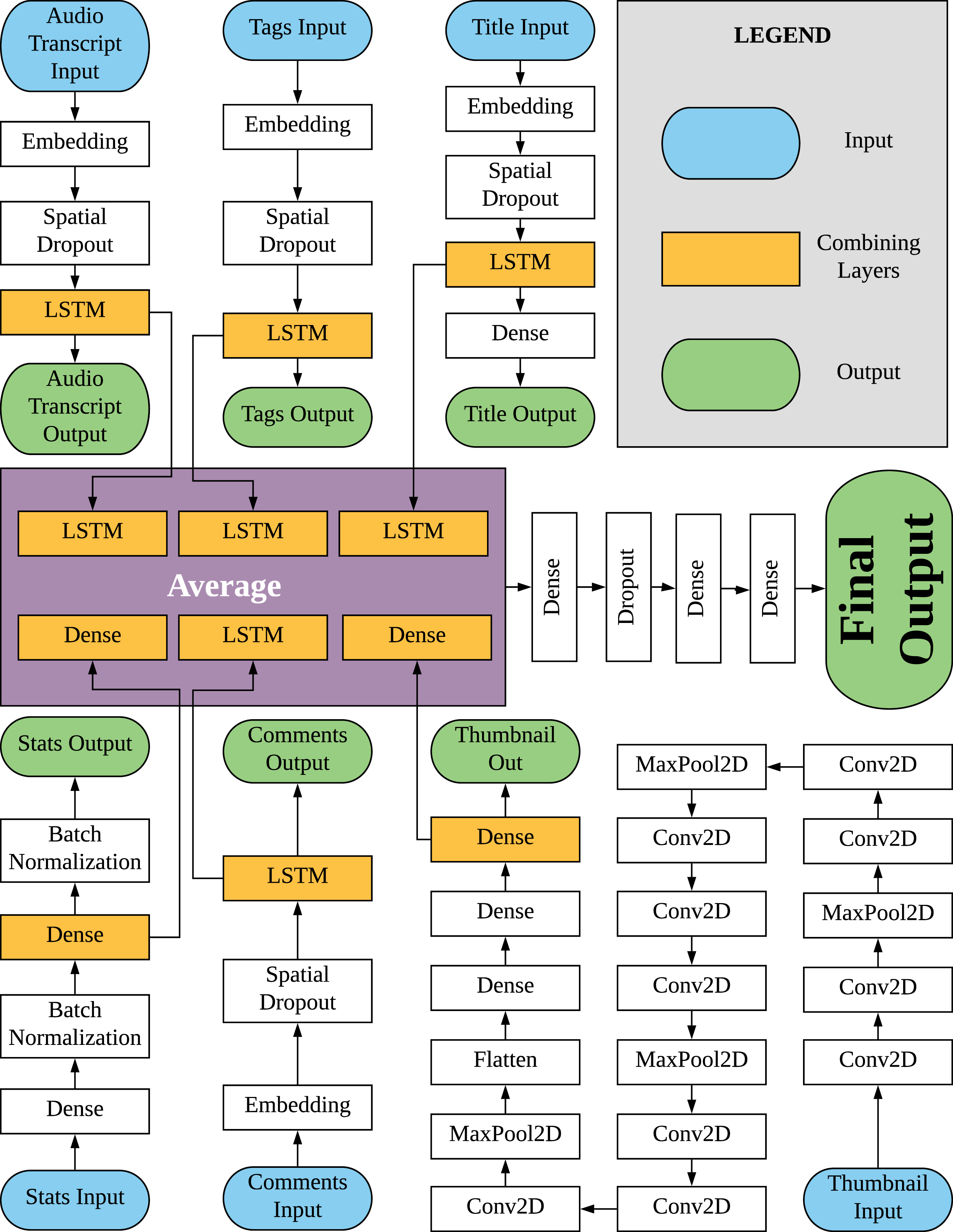}
\caption{Combined Model Architecture - \emph{BaitRadar}}
\label{fig::CombinedModel} \vspace{-5mm}
\end{figure}

Fig.~\ref{fig::CombinedModel} shows the final combined architecture of our model.
Instead of combining all the models into one, multiple tests are conducted to examine the effect of different combinations of the models with the Title model. 
Each model is trained by using the Early Stopping Approach, where the model automatically stops training when the training-loss reaches a certain threshold i.e., it converges. 
One advantage of this architecture is that it allows the system to function even when some inputs for the model are missing. 
This is a common problem on YouTube as comments may be disabled for some videos.
Finally, $X_{\text{final}}$ is computed as follows:
\begin{equation}
\label{Eq:average}
    X_\text{final} \leftarrow \frac{X_\text{t} + X_\text{tb} + X_\text{c} + X_\text{at} + X_\text{tg} + X_\text{s}}{n},
\end{equation}
where:
\begin{conditions}

 X_\text{final}     &  Average of all combined layers of individual models \\
 X_\text{t}     &  Output from LSTM cell of Title Model  \\
 X_\text{tb}    &  Output from Dense cell of Thumbnail Model  \\
 X_\text{c}     &  Output from LSTM cell of Comments Model  \\
 X_\text{at}    &  Output from LSTM cell of Audio Transcript Model  \\
 X_\text{tg}    &  Output from LSTM cell of Tags Model  \\
 X_\text{s}     &  Output from Dense cell of Statistics Model  \\
 n              &  Total number of individual models.
\end{conditions}

A keen-eyed reader can observe that the system will not be using the video frames to analyze the actual video, but instead, it uses the audio transcript.
The reason for this is trivial - imagine instead of a machine, a person is assigned the task of tagging clickbait videos within a given time constraint. 
The best approach would be to skim through different parts of the video and determine if it is clickbait. 
This can be implemented to a certain degree of accuracy, but when the decision is based on the semantic meaning of the video content, this approach becomes less effective. 
This is because still frames without any audio do not portray the actual video content, but rather its visual representation. 
Instead of the visual cue, the core meaning of the video can be obtained from the audio transcript, which computers can process faster since it is a text-based content. 

\begin{table}[t!]
\centering 
\caption{Accuracy of individual models\label{tab:accuracies}}
\small{
\begin{tabular}{|l|l|l|l|}
\hline
\textbf{Model}            & \textbf{Training } & \textbf{Validation} & \textbf{Testing} \\
& \textbf{Accuracy} & \textbf{Accuracy} & \textbf{Accuracy} \\
\hline
\textbf{Audio Transcript} & 99.09                      & 94.12                        & 93.80                      \\ \hline
\textbf{Title}            & 94.29                      & 87.20                         & 87.30                      \\ \hline
\textbf{Thumbnail}        & 90.61                      & 90.61                        & 81.43                     \\ \hline
\textbf{Comments}         & 98.64                      & 96.40                         & 96.80                      \\ \hline
\textbf{Tags}             & 99.83                      & 98.76                        & 98.70                      \\ \hline
\textbf{Statistics}       & 78.69                      & 78.69                        & 77.78                     \\ \hline
\end{tabular}}
\end{table}

\section{Experiment and Results}\label{sec::Experiments}
The proposed models are implemented on Tensorflow with Keras using Python and trained on Nvidia RTX 2080 Ti.
The performance of the proposed model is evaluated by using 
the dataset detailed in Section~\ref{sec::Proposed}.
First, the detection accuracy attained by each individual model is recorded in Table~\ref{tab:accuracies}.
Here, the accuracy is defined as 
(TP + TN)/(TP + TN + FP + FN), where TP refers to the true positive and the rest are defined in a similar manner. 
It is observed that the accuracy is $>77\%$ for all models.
Notably, the highest individual accuracy is achieved by the tags model.
A potential reason is because 
clickbait videos contain higher number of tags for tricking the YouTube recommendation algorithm.
On the other hand, the Statistics model yields the lowest accuracy. 
A potential reason is due to clickbait videos not having any outstanding statistics to easily identify them.

\begin{figure}[t!]
\centering
\includegraphics[width=\linewidth]{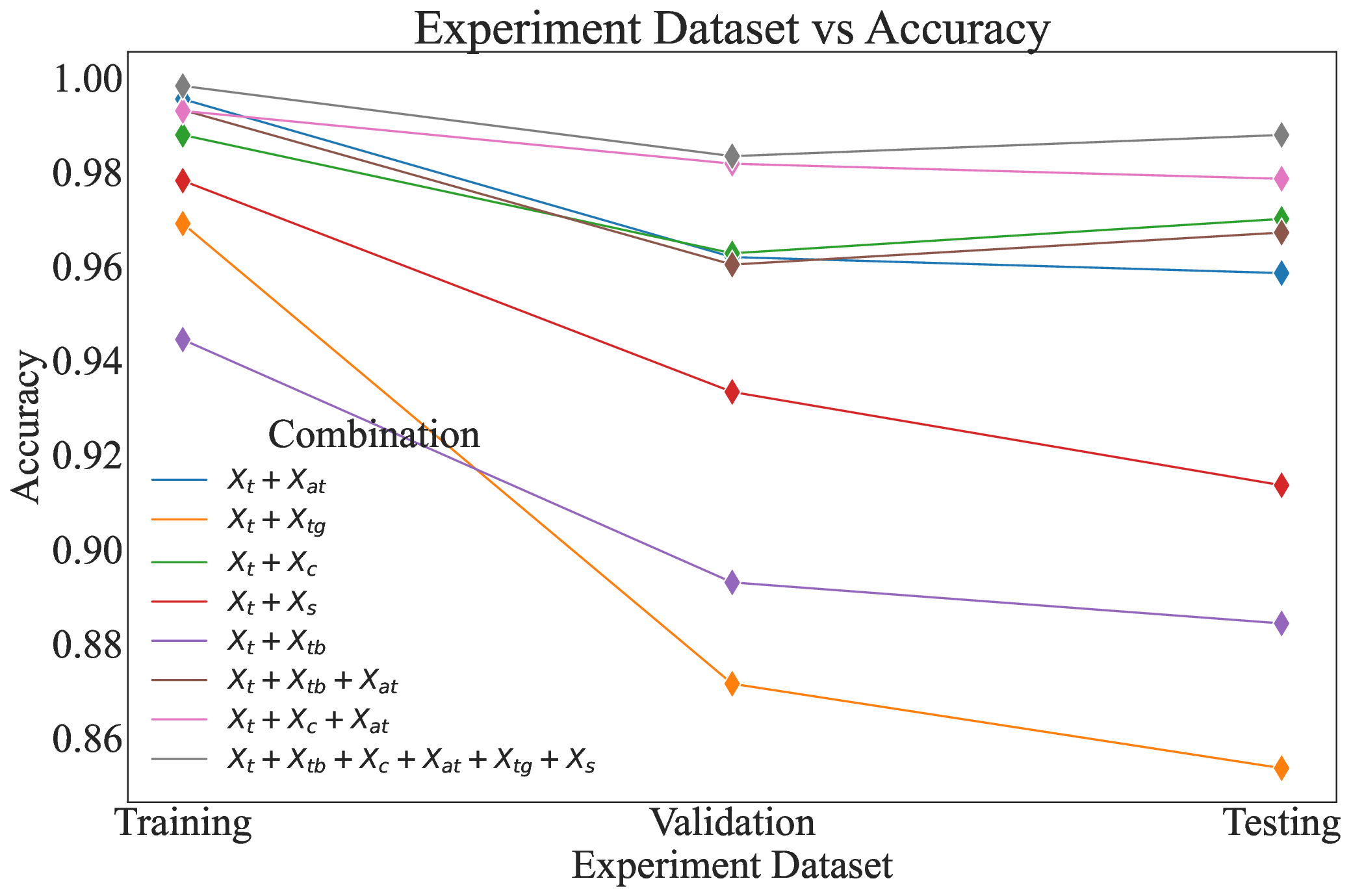}
\caption{Accuracies of different model combinations 
}
\label{fig::combinedgraph}
\end{figure}

Based on the results, the individual architectures with the highest accuracy are utilized to build the final model. 
Empirical testing was performed on different combinations of the models to determine which combination provides the highest accuracy.
Fig.~\ref{fig::combinedgraph} summarises the results obtained by different combinations of individual models. 
Results suggest that the combination of all six models provides the highest accuracy ($\sim98\%$) because it has the most data to train and infer with.
This agrees with intuition because for the case of a clickbait video, the actual content does not match the title as well as other attributes of the video. 
In other words, more data is available when more attributes are considered, hence higher chances to find any mismatches.
As expected, combining all models takes more epochs on average to converge in comparison to the other models because each (sub-)model is contributing to the overall accuracy.
For the rest of the discussion, this combination of models is referred to as \emph{BaitRadar}.

Next, experiments are carried out to verify the ability of BaitRadar to generalize. 
Specifically, BaitRadar is inferred on 100 random videos that are picked from YouTube channels while ensuring that they do not appear in any of the channels used in the YouTube dataset~\cite{gamage}.
Out of the 100 videos, 94 are classified correctly by BaitRadar, which translates to an accuracy of $94\%$, thereby confirming that BaitRadar can generalize the problem.

In order to verify that there is no bias due to the training methodology, 
we test whether the models are classifying based on the channel name and not learning the cues from the video.
For that purpose, non-clickbait videos from channels that are labelled as clickbait in the initial dataset are passed into the model. 
If the model is biased towards these channels, it would classify them as clickbait. 
However, after trying 20 non-clickbait videos from PewDiePie’s channel (which is classified as clickbait in the dataset), $>90\%$ of the time BaitRadar classifies the video as non-clickbait. 
This confirms that there is no bias in the training methodology. 

\section{Discussion}
\CH{As mentioned above, this work is inspired by \cite{Zannettou2018}, \WKS{which is} the only work that aims to solve clickbait on Youtube videos. 
Their model is able to achieve an accuracy of $92.4\%$, however a direct comparison cannot be carried out between the two datasets due to two main reasons: (1) the randomness in the selection of the videos from the Youtube channels and the exact list of videos are not provided by the researchers. Thus, the selected videos from each channel are different, and; (2) their model does not use the audio transcript of the video, which leads to an advantage for our model, thus not having a fair comparison.}

In comparison to classical machine learning, deep learning models can extract features automatically and make prediction based on these features. 
This enables deep learning models to learn features and patterns in the data that is not easily identifiable. 
Coupled with the large amounts of data generated by YouTube, BaitRadar is able to perform well. 
Using the training data, the model learns and updates the embedding vector, which is then utilized to tokenize the data. 
As an example in BaitRadar, the phrase ``mind blown" would carry a different weight when it appears in different attribute of the video (e.g., title v.s. audio transcript). 
The reason is that the occurrence of the phrase ``mind blown" in the title can indicate that the video is a clickbait, because a clickbait video tends to use provocative expressions to attract the users. 
In comparison, the phrase “mind blown” in the audio transcript model can indicate that the video is not clickbait as the creator in the video is talking about it, thereby the content matches the title.

Interestingly, the lowest accuracy was achieved when only the \emph{title} and \emph{tags} are combined. 
This is completely opposite of what was observed from the individual models where the tags model yields the highest accuracy. 
The potential reason is that the tags model starts over-fitting when it is combined with the title, because the training data set for both models are significantly smaller in comparison to the other models.
For example, the average title of a YouTube video contains $5 \sim 10$ words and the number of the tags in a video is $10 \sim 20$ words, which \WKS{are} short in comparison to comments or the audio transcript.
As a result, this combination of models fails to improve the accuracy to $>85\%$, hence ineffective.

Since clickbait detection is a subjective topic, our model has some issues with borderline cases, i.e., videos that are not clearly considered as clickbait.
For example, \cite{Meloche_2019} is a simple vlog. 
From the perspective of a subscriber of this channel, this video is not a clickbait, since its purpose is to entertain the loyal subscribers. 
However, \WKS{when} a person who is new to this channel comes across this video, then the thumbnail of this video is a clickbait to him/her. 
In this case, our model makes an inference based on the opinion of a non-subscriber (i.e., clickbait). 
This is a limitation of our model, which will be addressed as future work.
Nonetheless, considering the results and analysis obtained from the deep learning models, BaitRadar is performing as intended.

\section{Conclusions\label{sec::Conclusion}}
\CH{In this work}, the clickbait problem on YouTube is identified and a model is proposed to classify clickbait videos.
The proposed model utilizes deep learning techniques to classify clickbait videos by using the Title, Thumbnail, Comments, Audio Transcript, Tags and other statistics of the video. 
A custom architecture is proposed to extract and learn the features from each video attribute and then the individual models are combined to create the multi-model architecture.
Different combinations of models are trained and explored to understand the effectiveness of each video attribute, and the best combination \WKS{is} adopted as the BaitRadar model.
This model achieves an accuracy of $98\%$ with an inference time of $\leq2s$. \BG{As future work, we will explore the performance of the model on various kinds of media and aim to also understand the different insights from the model interpretability to better detect clickbait videos.}
%
%
%
%
\bibliographystyle{IEEEbib}
\bibliography{refs}




    
    
    




\end{document}